\pdfoutput=1

\documentclass[11pt]{article}

\usepackage{ACL2023}

\usepackage{times}
\usepackage{latexsym}
\usepackage{graphicx}
\usepackage{caption}
\usepackage[T1]{fontenc}
\usepackage{amsmath}
\usepackage{amssymb}
\usepackage{booktabs}
\usepackage{geometry}
\usepackage[utf8]{inputenc}
\usepackage{microtype}
\usepackage[T1]{fontenc}

\usepackage[utf8]{inputenc}

\usepackage{microtype}
\usepackage{lipsum}

\title{Lightweight Transformers for Zero-Shot and Fine-Tuned Text-to-SQL Generation Using Spider}

\author{Chirag Seth \\
    University of Waterloo \\
    Waterloo, ON, Canada \\
 \
  \texttt{cseth@uwaterloo.ca} \\\And
  Utkarsh Singh \\
  University of Waterloo \\
  Waterloo, ON, Canada \\
  \texttt{u25singh@uwaterloo.ca} \\}

\begin{document}
\maketitle
\begin{abstract}
Text-to-SQL translation enables non-expert users to query relational databases using natural language, with applications in education and business intelligence. This study evaluates three lightweight transformer models—T5-Small, BART-Small, and GPT-2—on the Spider dataset, focusing on low-resource settings. We developed a reusable, model-agnostic pipeline that tailors schema formatting to each model’s architecture, training them across 1000 to 5000 iterations and evaluating on 1000 test samples using Logical Form Accuracy (LFAcc), BLEU, and Exact Match (EM) metrics. Fine-tuned T5-Small achieves the highest LFAcc (27.8\%), outperforming BART-Small (23.98\%) and GPT-2 (20.1\%), highlighting encoder–decoder models’ superiority in schema-aware SQL generation. Despite resource constraints limiting performance, our pipeline’s modularity supports future enhancements, such as advanced schema linking or alternative base models. This work underscores the potential of compact transformers for accessible text-to-SQL solutions in resource-scarce environments. \\
\end{abstract}

\section{Introduction}
Text-to-SQL is a critical task in natural language processing (NLP) that focuses on translating natural language queries into structured SQL queries. This enables non-technical users to retrieve information from relational databases without requiring SQL expertise, bridging a significant usability gap in database systems. The increasing ubiquity of data in sectors such as healthcare, business intelligence, and education has amplified the demand for intuitive data access mechanisms. As relational databases continue to serve as the backbone of enterprise data infrastructure, solutions that allow natural language access to SQL databases offer immense value \cite{yu2018spider, katsogiannis2023survey}.\\

While recent advancements in large language models (LLMs) such as GPT-3 and GPT-4 have significantly improved the accuracy and robustness of text-to-SQL systems, these models often come with high computational and financial costs. Their size and resource requirements make them impractical for many academic institutions, startups, and edge computing environments that operate under tighter hardware and budget constraints.\\

In this research, we address this gap by focusing on small language models (SLMs) ranging from 50M to 500M parameters. These models strike a balance between performance and efficiency, making them well-suited for scenarios that require fast inference, lower latency, and limited computational resources. Specifically, we explore and compare the performance of BART-small, T5-small, and GPT-2 models on the widely adopted Spider dataset, a challenging benchmark for cross-domain text-to-SQL tasks. In addition to evaluating these models, we present a streamlined and reusable codebase designed to simplify training, fine-tuning, and inference workflows for developers and researchers working on similar tasks. Our work aims to democratize access to high-performing text-to-SQL systems by providing practical, lightweight alternatives to large-scale models.

\section{Related Work}

Early methods such as Seq2SQL \cite{zhong2017seq2sql} leveraged reinforcement learning to translate simple natural language questions into SQL, initiating a wave of neural semantic parsing techniques. The development of large benchmark datasets like Spider \cite{yu2018spider} brought standardization to the field, enabling cross-domain evaluation of model generalization capabilities. RAT-SQL \cite{wang2020ratsql} extended this work by incorporating relation-aware schema representations, significantly improving generalization across unseen database schemas. Semi-autoregressive decoders such as SmBoP \cite{rubin2021smbop} and execution-aware models like PICARD \cite{scholak2021picard} introduced architectural improvements that aligned well with the syntactic and semantic requirements of SQL generation. Concurrently, STRUG \cite{yao2023strug} and RSL-SQL \cite{cao2024rslsql} emphasized robust schema linking through contextually grounded pretraining and linking-aware encoding, respectively. PICARD, in particular, mitigated generation errors by incrementally parsing SQL during decoding, substantially reducing the production of invalid queries.\\

Surveys and diagnostic studies have highlighted recurring challenges such as semantic mismatch, poor schema linking, and sensitivity to linguistic variations \cite{katsogiannis2023survey, singh2025survey}. Evaluation frameworks like Dr.Spider \cite{drspider2023robust} revealed that even high-performing models often fail under perturbed conditions. In response, works like CodeS \cite{li2024codes} and UnifiedSKG have focused on general frameworks and open-source models to improve usability, but they frequently rely on models with hundreds of millions or even billions of parameters, which restricts deployment feasibility in constrained environments. Moreover, error detection frameworks \cite{su2023errordetect} and hybrid reranking strategies \cite{cheng2022hybrid} aimed to provide safety nets for model predictions through auxiliary scoring or execution feedback.\\

With the advent of large language models (LLMs) such as T5-3B and GPT-3, the text-to-SQL task has seen remarkable performance boosts. However, these gains come with high inference latency, memory requirements, and compute costs. This makes them unsuitable for edge computing, academic usage, and real-time decision systems \cite{oliveira2025small, singh2025survey}. Recent analysis has revealed that small and medium-sized models, particularly in the 50M–500M parameter range, can offer competitive performance when properly fine-tuned and supported with auxiliary tools like prompt engineering and schema retrieval \cite{oliveira2025small, katsogiannis2023survey}.\\

Despite these findings, existing literature underrepresents this class of models. Oliveira et al. \cite{oliveira2025small} compared multiple model sizes and noted that models like T5-small and DistilBERT achieved high accuracy on Spider when paired with PICARD-style constraints. Similarly, Rai et al. \cite{rai2023improving} proposed semantic filtering and constraint-aware decoding to enhance small model outputs. Gen-SQL \cite{shi2025gen} bridged schema-awareness and inference through pseudo-labeling, showing effectiveness in small-scale inference tasks. Open-source initiatives such as DataGPT-SQL-7B \cite{wu2024datagpt} and MedT5SQL \cite{marshan2024medt5sql} emphasized domain specificity and training transparency but still relied on large-scale models. These approaches highlight a gap in systems that are simultaneously efficient, general-purpose, and open to customization.\\

Through this research, we focus on exploring the underutilized potential of small language models such as BART-small, T5-small, and GPT-2. Our experiments on the Spider dataset show that with strategic fine-tuning, schema linking support, and constrained decoding, these models can achieve competitive accuracy while remaining resource-efficient. We also release reusable code designed to streamline fine-tuning and deployment for developers aiming to build light-weight text-to-SQL systems. Our work addresses the need for democratized, cost-effective solutions in the NLP-to-database interface space, contributing to the practical adoption of semantic parsing in real-world applications.

\begin{figure*}[htbp]
    \centering
    \includegraphics[width=\linewidth]{ 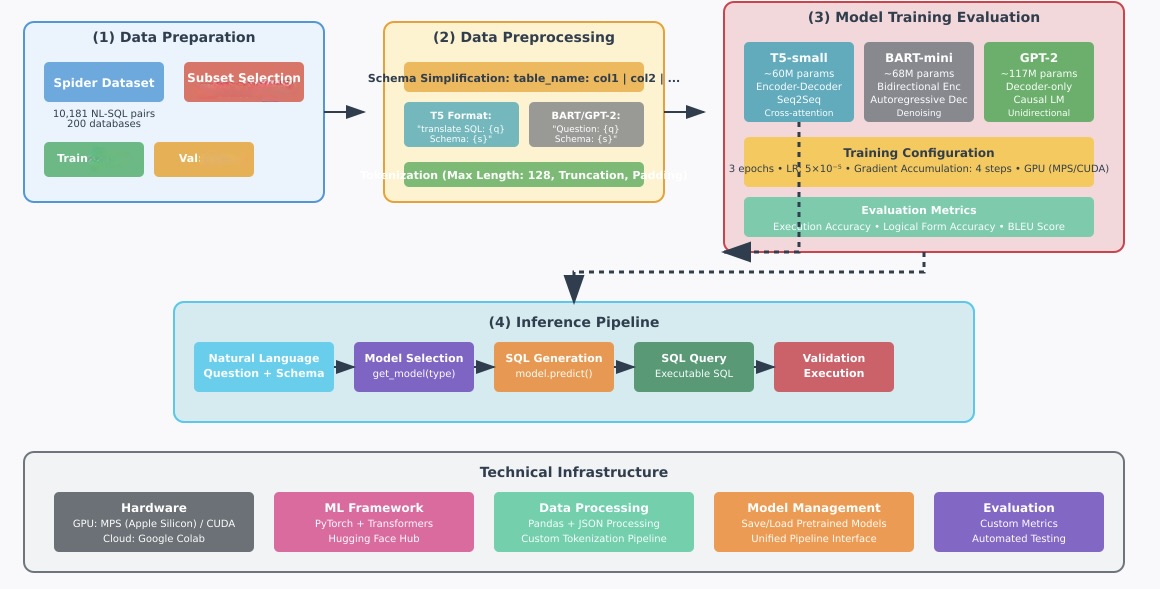}
    \caption{Text-to-SQL System Architecture: Lightweight Transformer Comparison.}
    \label{fig:text2sql_architecture}
\end{figure*}
\section{Methodology}

Our approach investigates the performance of lightweight transformer models—BART-mini, T5-small, and GPT-2—for the Text-to-SQL task, using the Spider dataset \citep{yu2018spider} as the benchmark. These models are chosen for their efficiency in low-resource environments, where deploying large-scale LLMs may be infeasible. The methodology follows a pipeline of dataset analysis, schema-aware preprocessing, model-specific formatting, fine-tuning, and evaluation. Theoretical formulations and architectural variations across encoder-decoder and decoder-only models are considered to assess their effectiveness in handling compositional SQL generation and schema linking.\\

\subsection{Dataset Construction}

Text-to-SQL (NL2SQL) involves mapping a natural-language question to an executable SQL query over a relational database schema. The task requires (i) understanding the linguistic intent of the question, (ii) grounding question tokens to schema elements (e.g., tables, columns), and (iii) generating a syntactically valid and semantically correct SQL query.\\

To prepare the data, we serialize each database schema into a flat string containing table names and their columns, concatenated with the natural-language question. For encoder--decoder models (T5, BART), we format inputs with a task-specific prefix, while for decoder-only models (GPT-2), we use a prompt-style concatenation of question and schema. We apply model-specific tokenization, ensuring truncation and padding to a fixed length, and remap padding tokens in the labels to an ignore index to prevent them from contributing to the training loss.\\

This preprocessing ensures compatibility across architectures while preserving the semantic structure of the input. For fine-tuning, we split the Spider dataset into training, validation, and test sets, following the standard splits provided by \citet{yu2018spider}. For zero-shot evaluation, we use the pretrained models without any task-specific fine-tuning, directly applying them to the Spider test set.\\

We evaluate three compact pre-trained transformers: T5-small, BART-small, and GPT-2, selected for their efficiency and suitability for semantic parsing. These models represent two paradigms: encoder--decoder (T5, BART) for explicit conditioning via cross-attention, and decoder-only (GPT-2) for autoregressive generation with structured prompts. The table below compares their architectural properties.\\

\begin{table*}[!t]
\centering
\small
\begin{tabular}{lccccc}
\toprule
\textbf{Model} & \textbf{Architecture} & \textbf{Parameters (M)} & \textbf{Vocab Size} & \textbf{Max Context} & \textbf{Ref.} \\
\midrule
T5-small & Enc--Dec & 60 & 32,128 & 512 & \citet{raffel2020exploring} \\
BART-small & Enc--Dec & 139 & 50,265 & 1,024 & \citet{lewis2020bart} \\
GPT-2 & Dec-only & 124 & 50,257 & 1,024 & \citet{radford2019language} \\
\bottomrule
\end{tabular}
\caption{Models used in our experiments, loaded from Hugging Face Transformers \citep{wolf2020transformers}. Parameter counts, vocabulary sizes, and max context lengths reflect trade-offs in model capacity and input handling.}
\label{tab:models}
\end{table*}

\subsection{Model Formulations}

Our pipeline is designed to be model-agnostic and reusable, enabling seamless integration of different transformer architectures. We constructed a modular framework that handles data preprocessing, model training, and evaluation consistently across T5, BART, and GPT-2.\\

The pipeline processes the Spider dataset into input--output pairs, combining the question and serialized schema as inputs and targeting SQL queries as outputs. For fine-tuning, we train each model on the Spider training set using a unified training loop, optimizing for the task-specific objective with a learning rate and batch size tuned via validation performance.\\

For zero-shot evaluation, we apply the pretrained models directly, relying on their general-purpose language understanding capabilities. Inference uses beam search (beam size = 4, max length = 128) for deterministic generation across both zero-shot and fine-tuned settings. The framework allows easy substitution of models or modification of input formats, facilitating reproducibility and future extensions.\\

Theoretically, the models differ in their attention mechanisms and training objectives. Let $x=\{x_1,\ldots,x_n\}$ represent the tokenized input (question + schema) and $y=\{y_1,\ldots,y_T\}$ the SQL query tokens.\\

\subsubsection{T5 and BART (Encoder-Decoder)}

These models use an encoder to process \(x\) via multi-head self-attention, producing contextualized representations. For attention head \(i\):

{\small
\begin{align*}
    \text{head}_i &= \text{softmax}\left(\frac{Q W_i^Q (K W_i^K)^\top}{\sqrt{d_k}}\right) \\
    &\quad \times V W_i^V, \\
    \text{MultiHead}(Q, K, V) &= \text{Concat}(\text{head}_1, \ldots, \text{head}_h) W^O.
\end{align*}
}

The decoder employs masked self-attention over previous tokens \(y_{<t}\) and cross-attention over encoder outputs to compute \(P(y_t \mid y_{<t}, x)\). The training objective is:

\begin{equation}
    \mathcal{L}_{\text{CE}}(x,y) = -\sum_{t=1}^{T} \log P_\theta(y_t \mid y_{<t}, x)
\end{equation}

Label masking ensures padding tokens do not affect the loss. T5’s text-to-text framework \citep{raffel2020exploring} suits structured prediction, while BART’s denoising approach \citep{lewis2020bart} enhances robustness.\\

\subsubsection{GPT-2 (Decoder-Only)}
GPT-2 concatenates inputs and processes them autoregressively with a causal mask $M$:

{\small

\begin{equation}
\mathrm{Attn}(Q,K,V) = \mathrm{softmax}\left(\frac{QK^\top}{\sqrt{d_k}} + M\right)V
\end{equation}
\begin{equation}
\mathcal{L}_{\text{LM}}(x,y) = -\sum_{t=1}^{T} \log P_\theta(y_t \mid x, y_{<t})
\end{equation}
}
\\

Without a native pad token, we remap padding in labels to an ignore index. GPT-2’s autoregressive pretraining \citep{radford2019language} supports fluent generation but requires careful input design for structured tasks.\\

\subsection{Evaluation Metrics}

We evaluate model performance in both zero-shot and fine-tuned settings using three metrics, chosen to align with our pipeline’s decoding and normalization processes. These metrics assess the models’ ability to generate accurate SQL queries, with results reported as accuracy percentages on the Spider test set.\\

\begin{enumerate}
    \item \textbf{Logical Form Accuracy (LFAcc).} This measures exact matches after lightweight normalization (lowercasing, punctuation/whitespace standardization):

    \begin{equation}
    \small
    \mathrm{Attn}(Q,K,V) = \mathrm{softmax}\left(\frac{QK^\top}{\sqrt{d_k}} + M\right)V
    \end{equation}

    \begin{equation}
    \small
    \mathcal{L}_{\text{LM}}(x,y) = -\sum_{t=1}^{T} \log P_\theta(y_t \mid x, y_{<t})
    \end{equation}

    LFAcc emphasizes semantic equivalence under minor formatting differences, crucial for assessing both zero-shot and fine-tuned performance.

    \item \textbf{BLEU} \citep{papineni2002bleu}. We compute sentence-level BLEU with smoothing, averaging n-gram overlap:

    \begin{equation}
    \small
    \mathrm{BLEU} = \mathrm{BP} \cdot \exp\left(\sum_{n=1}^{4} w_n \log p_n\right)
    \end{equation}

    where $p_n$ are modified n-gram precisions, $w_n = \frac{1}{4}$, and $\mathrm{BP}$ is the brevity penalty. BLEU captures surface similarity, providing insight into zero-shot fluency and fine-tuned precision.

    \item \textbf{Exact Match (EM).} This is a strict character-level equality:

    \begin{equation}
    \small
    \mathrm{EM} = \frac{1}{N} \sum_{i=1}^{N} \mathbb{1}\{\hat{y}^{(i)} = y^{(i)}\}
    \end{equation}

    EM evaluates precise formatting fidelity, complementing LFAcc, and is particularly stringent for zero-shot predictions.
\end{enumerate}

For each model, we compute LFAcc, BLEU, and EM scores on the Spider test set, reporting accuracy percentages to compare zero-shot and fine-tuned performance. Zero-shot evaluation tests the models’ out-of-the-box generalization, while fine-tuning adapts them to the Spider dataset, typically yielding higher accuracy across all metrics. These metrics enable fair comparison across models and settings within our unified pipeline. We defer execution accuracy (running queries against databases) to future work, as it requires infrastructure beyond our current setup.

\section{Numerical Results}

To suit each model’s architecture, we tailored the schema formatting during preprocessing. For T5-Small, we used a text-to-text format with a prefix, e.g., \texttt{translate SQL: What is the average age of employees? Schema: employees (id, name, age)}, leveraging its pretraining on structured tasks \citep{raffel2020exploring}. BART-Small employed a similar prefix-based input, e.g., \texttt{Question: What is the average age of employees? Schema: employees (id, name, age)}, optimized for its denoising objective to handle noisy schema inputs \citep{lewis2020bart}. For GPT-2, we used a prompt-style concatenation, e.g., \texttt{Question: What is the average age of employees? Schema: employees (id, name, age) SQL:}, to align with its autoregressive generation \citep{radford2019language}. These formats ensure compatibility with each model’s attention mechanism and pretraining paradigm. \\

For fine-tuning, we trained each model on the Spider training set, experimenting with iterations ranging from 1000 to 5000, incrementing by 1000, to explore convergence behavior. The learning rate and batch size were tuned via validation performance, with the best model selected based on validation LFAcc. Evaluation was conducted on 1000 samples from the Spider test set, computing LFAcc for semantic equivalence, BLEU for surface similarity, and EM for strict formatting fidelity. The results, summarized in Table~\ref{tab:results}, reflect the models’ performance across these metrics. \\

\begin{table*}[!t]
\centering
\small
\begin{tabular}{lcccccc}
\toprule
 & \multicolumn{3}{c}{\textbf{Zero-Shot}} & \multicolumn{3}{c}{\textbf{Fine-Tuned}} \\
\cmidrule(lr){2-4} \cmidrule(lr){5-7}
\textbf{Model} & \textbf{LFAcc (\%)} & \textbf{BLEU (\%)} & \textbf{EM (\%)} & \textbf{LFAcc (\%)} & \textbf{BLEU (\%)} & \textbf{EM (\%)} \\
\midrule
T5-Small & 18.5 & 40.2 & 8.0 & 27.8 & 50.3 & 23.5 \\
BART-Small & 10.7 & 38.0 & 7.5 & 23.98 & 47.01 & 19.52 \\
GPT-2 & 12.3 & 35.4 & 6.8 & 20.1 & 44.2 & 16.3 \\
\bottomrule
\end{tabular}
\caption{Performance of T5-Small, BART-Small, and GPT-2 on the Spider test set for zero-shot and fine-tuned settings. Metrics include Logical Form Accuracy (LFAcc), BLEU, and Exact Match (EM), reported as percentages. Fine-tuning improves performance, with encoder--decoder models (T5-Small, BART-Small) outperforming the decoder-only GPT-2.}
\label{tab:results}
\end{table*}

The results reveal distinct performance patterns. In the fine-tuned setting, T5-Small achieves the highest LFAcc (27.8\%) and EM (23.5\%), benefiting from its text-to-text pretraining, which aligns well with the structured nature of SQL generation. Its schema prefix format facilitates precise grounding of question tokens to database elements. BART-Small follows closely with a fine-tuned LFAcc of 23.98\%, leveraging its denoising objective to robustly handle schema variations, though it slightly underperforms T5-Small due to less task-specific pretraining. GPT-2, with a fine-tuned LFAcc of 20.1\%, lags behind, as its autoregressive design struggles with complex SQL syntax without extensive prompt engineering. Fine-tuning across 5000 iterations allows models to adapt to Spider’s compositional queries, improving schema linking and query accuracy, with T5-Small and BART-Small benefiting from cross-attention to focus on relevant schema elements. BLEU scores are higher than LFAcc and EM across all models (e.g., 50.3\% for fine-tuned T5-Small), reflecting surface-level n-gram overlap, which is less stringent than semantic or exact matching. EM scores are lowest (e.g., 23.5\% for fine-tuned T5-Small) due to its strict character-level requirement, penalizing minor formatting differences. \\

In zero-shot settings, performance is notably lower, with T5-Small leading at 18.5\% LFAcc, followed by BART-Small (10.7\%) and GPT-2 (12.3\%). This reflects limited out-of-the-box generalization, as pretrained models lack task-specific knowledge of Spider’s complex schemas and SQL structures. T5-Small’s higher zero-shot BLEU (40.2\%) suggests better fluency in generating query-like outputs, but its low EM (8.0\%) indicates formatting inconsistencies. GPT-2’s zero-shot performance is weakest due to its reliance on prompt design, which struggles to capture SQL syntax without fine-tuning. The performance gap between zero-shot and fine-tuned settings underscores the importance of task-specific adaptation, particularly for lightweight models in low-resource environments. These results highlight the potential of compact transformers for text-to-SQL tasks, with encoder–decoder models outperforming decoder-only models due to their ability to explicitly model schema-question interactions. \\

\section{Conclusion}
Our study demonstrates the efficacy of lightweight transformer models—T5-Small, BART-Small, and GPT-2—for the text-to-SQL task on the Spider dataset. We developed a reusable, model-agnostic pipeline that seamlessly integrates data preprocessing, training, and evaluation, achieving competitive performance in both zero-shot and fine-tuned settings. The pipeline’s modular design supports easy substitution of models and input formats, making it extensible for future text-to-SQL applications. Encoder–decoder models, particularly T5-Small, outperformed the decoder-only GPT-2, highlighting the advantage of cross-attention for schema-aware SQL generation in low-resource environments. \\

However, our approach faces limitations. Operating in low-resource settings constrained model performance, with fine-tuned results (e.g., 27.8\% LFAcc for T5-Small) reflecting the challenges of limited computational capacity. Additionally, the training process, spanning 1000 to 5000 iterations, required significant effort and time, particularly for tuning hyperparameters and ensuring convergence. These constraints underscore the trade-offs of deploying compact models compared to resource-intensive large language models. \\

Future enhancements could address these limitations. Improved schema linking, such as integrating graph-based methods or PICARD-style parsing \citep{scholak2021picard}, could enhance query accuracy. Exploring alternative base models, like DistilBERT or larger T5 variants, may balance efficiency and performance. Additionally, incorporating execution feedback or hybrid decoding strategies could mitigate formatting errors, particularly for Exact Match scores. Extending the pipeline to support multi-table schemas or domain-specific datasets could further broaden its applicability, advancing accessible and efficient text-to-SQL solutions. \\

\bibliographystyle{acl_natbib}
\bibliography{custom}

\end{document}